\documentclass[10pt, conference, compsocconf]{IEEEtran}
\renewcommand{\IEEEbibitemsep}{1pt plus 2pt}
\makeatletter
\IEEEtriggercmd{\reset@font\normalfont\footnotesize}
\makeatother
\IEEEtriggeratref{1}

\usepackage[cmex10]{amsmath}
\usepackage{amssymb}
\usepackage{booktabs}
\usepackage{tikz}

\usepackage{graphicx}
\usepackage[colorinlistoftodos]{todonotes}
\usepackage[colorlinks=true, allcolors=blue,bookmarks=false]{hyperref}
\usepackage{breqn}
\usepackage{subcaption}
\usepackage{array}
\newcolumntype{C}[1]{>{\centering\arraybackslash}p{#1}}

\usepackage{cite}

\ifCLASSINFOpdf
\else
\fi

\usepackage{stfloats}
\begin{document}
%
\title{Direct Fitting of Gaussian Mixture Models}


\author{\IEEEauthorblockN{Leonid Keselman,
Martial Hebert}
\IEEEauthorblockA{Robotics Institute\\
Carnegie Mellon University\\
Pittsburgh, PA, USA\\
leonidk@cmu.edu,
hebert@ri.cmu.edu}
}

\maketitle

\begin{abstract}
When fitting Gaussian Mixture Models to 3D geometry, the model is typically fit to point clouds, even when the shapes were obtained as 3D meshes. Here we present a formulation for fitting Gaussian Mixture Models (GMMs) directly to a triangular mesh instead of using points sampled from its surface. Part of this work analyzes a general formulation for evaluating likelihood of geometric objects. This modification enables fitting higher-quality GMMs under a wider range of initialization conditions. Additionally, models obtained from this fitting method are shown to produce an improvement in 3D registration for both meshes and RGB-D frames. This result is general and applicable to arbitrary geometric objects, including  representing uncertainty from sensor measurements. 
\end{abstract}


%
\IEEEpeerreviewmaketitle

\section{Introduction}\label{sec:intro}
In robotics and computer vision, there exist many forms of representation for 3D geometric data. For example, some researchers use unordered point sets~\cite{mckaybesl}, others require points with surface normals~\cite{SGP:SGP06:061-070} or dense volumetric representations such as signed distance fields~\cite{KinectFusion}. The variation in forms of representation is related to the wide variety of sources and uses for this data, from raw depth sensor measurements to Computed-Aided Design (CAD) models. 

\begin{figure}
  \centering
    \includegraphics[width=\linewidth]{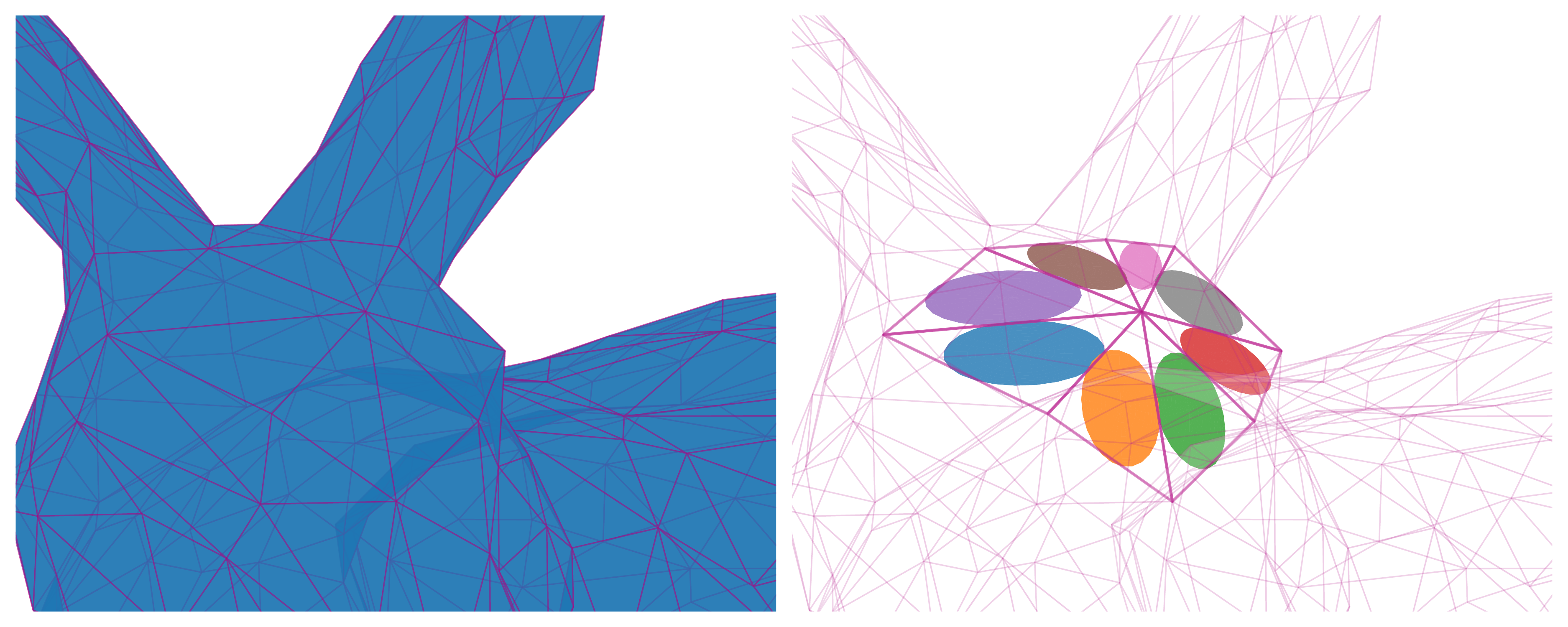}
    \caption{\textbf{Visual example of the Stanford Bunny}, highlighting 8 triangles on the head. Each triangle is shown with its covariance (plotted to $1.5 \sigma$). We demonstrate how Gaussian Mixture Models can use the covariance information from given geometric structures (in this case, triangles) to fit models more efficiently, robustly and with higher fidelity. }
  \label{fig:head}
\end{figure}

Many researchers have found that Gaussian Mixture Models provide a powerful representation, especially for use in registration between unknown poses~\cite{Jian2011,Eckart2015,Eckart2016,Eckart2018,wenniegmm}. Producing a Gaussian Mixture Model requires only unstructured point sets from the underlying geometric data, which makes them widely applicable. Our contribution in this work is to demonstrate how Gaussian Mixture Model can be constructed, evaluated on and fit directly to geometric primitives, such as the triangles of a polygon mesh. This is done by incorporating the structural information from each primitive into the algorithm, for a visual example, see Figure~\ref{fig:head}. 

Our contributions include a mathematical framework for how geometric primitives can be incorporated with probability distributions (Section~\ref{sec:prodint}). We demonstrate how to obtain the structural properties for a triangular mesh (Section~\ref{sec:method}) and how it can be generalized to other primitives (Section~\ref{sec:general}).
Incorporating structural information allows us to build Gaussian Mixture Models that not only converge faster and in more conditions (Section~\ref{sec:results}) but also provide a representation that produces higher quality registration results when the models are used in practice (Section~\ref{sec:applications}).

The code for all methods and experiments in this paper is available at \url{https://github.com/leonidk/direct_gmm}. 

\section{Method}\label{sec:method}

\subsection{Gaussian Mixture Models}
The Gaussian Mixture Model (GMMs) is a well studied probability distribution. It is possible to fit these models to empirical data via Maximum Likelihood Estimation (MLE)~\cite{Hasselblad1966,Dempster1977}. The likelihood of any point $x$ in a Gaussian is given by 
\begin{equation}
\mathcal{N}(x; \mu, \Sigma) = 2 \pi^{-k/2} \det( \Sigma)^{-\frac{1}{2}} e^{-\frac{1}{2} (x-\mu)^T \Sigma^{-1} (x-\mu)}
\label{eq:pdf}
\end{equation}
where $\mu \in \mathbb{R}^k$ is the mean and $\Sigma \in \mathbb{R}^{k \times k}$ is the positive-definite covariance matrix. The log-likelihood of a given point $x$ is given by
\begin{equation} 
\begin{aligned}\log \mathcal{N}(x; \mu, \Sigma) = &-\frac{k}{2}\log(2 \pi)\\
&- \frac{1}{2} \log(\det(\Sigma))\\
&-\frac{1}{2} (x-\mu)^T \Sigma^{-1} (x-\mu)
\end{aligned}
\label{eq:llpdf}
\end{equation}

In a Gaussian Mixture Model with $K$ components, with $\lambda_i$ are mixing weights subject to $\sum_i \lambda_i = 1$ and  $ \lambda_i \geq 0\ \forall i$, the probability of a point $x$ is given by
$$g(x) = \sum_{i=1}^{K} \lambda_i\ \mathcal{N}(x;\mu_i, \Sigma_i)$$

\subsection{Geometric Objects in a Probability Distribution}\label{sec:prodint}
First we must handle the general case of how to evaluate the probability of a known geometric object in a probability distribution. Consider sampling N points from the geometric object, where a notion of likelihood can be evaluated by taking their product. To account for the variable number of samples, we take a geometric mean. And to obtain the likelihood of the object, we take the limit as the number of samples grows to infinity. 
\begin{equation} \label{eq:gmm2}
\begin{split}
\ell &\cong \prod_{i = 1}^{N}{p(x_{i})} \\
\ell &\cong \left( \prod_{i = 1}^{N}{p(x_{i})} \right)^{\frac{1}{N}}\\
\ell &= \lim_{N\to\infty} \left( \prod_{i = 1}^{N}{p(x_{i})} \right)^{\frac{1}{N}}\\
\ell &= \lim_{N\to\infty} \exp \left(\log\left( \left( \prod_{i = 1}^{N}{p(x_{i})} \right)^{\frac{1}{N}}\right)\right)\\
\ell &= \lim_{N\to\infty} \exp \left(\frac{1}{N}  \sum_{i = 1}^{N}{\log (p(x_{i}))} \right)\\
\ell &= \exp \left( \int {\log( p(x))} dx \right) 
\end{split}
\end{equation}

Equation~\ref{eq:gmm2} is a form of product integral~\cite{feynman1951operator,guenther1983product,BASHIROV200836}, which can be used to evaluate the likelihood of a known geometric object. The extension to multiple objects is straightforward, simply adjust their sampling weights accordingly. This form of product integral has two nice properties, it is invariant to resampling and it produces a result of $0$ if $p(x)=0$ anywhere along the geometric object. By invariance to resampling, we mean that one large primitive with sophisticated integration bounds gives the same answer as many small disjoint pieces of surface with their own bounds. 

Following the use of Jensen's inequality, we get the following lower bound on likelihood
\begin{equation}
\begin{split}
L &=  \exp \left( \sum_{j=1}^{M} \int_\triangle \log \left( \sum_{i=1}^{K} \lambda_i\ \mathcal{N}(x;\mu_i, \Sigma_i)\right) dx \right) \\ 
\log(L) &=  \sum_{j=1}^{M} \int_\triangle \log \left( \sum_{i=1}^{K} \lambda_i\ \mathcal{N}(x;\mu_i, \Sigma_i)\right) dx \\
&\geq \sum_{j=1}^{M}  \sum_{i=1}^{K} \int_\triangle \log \left( \lambda_i\  \mathcal{N}(x;\mu_i, \Sigma_i)\right)dx
\label{eq:prodint}
\end{split}
\end{equation}

We also consider a simplified model, where each triangle is sampled at its center of mass ($\mu_j$), and has weight corresponding to its area ($\alpha_j$). As combining probabilities is done with multiplication, we use a weighted geometric mean over all points, obtaining the following approximation
\begin{equation}
L \approx L_S  = \prod_{j=1}^{M} \left( \sum_{i=1}^{K}  \lambda_i\ \mathcal{N}(\mu_j;\mu_i, \Sigma_i)\right)^{\frac{\alpha_j}{\sum_k \alpha_{k}}}
\label{eq:approx1}
\end{equation}

\section{Modifying EM maximization to account for triangles}
In this section we derive the expressions for fitting a GMM to a triangular mesh. We will represent each GMM component with parameters $\mu_i, \Sigma_i,\lambda_i$ and each triangle with vertices $A_j,B_j,C_j$, centroid $\mu_j$ and area $\alpha_j$. Traditional EM minimization is possible by analyzing the lower bound of the log-likelihood. Following standard formulations (see \cite{Dempster1977,jordan2009,sridharan2014gaussian}), we add mixture sampling probabilities $\eta_{ij}$ and move the logarithm inside the summation to obtain a valid lower-bound to minimize by using Jensen's inequality. 

To perform fitting, we need an M-step which obtains $\lambda_i,\mu_i,\Sigma_i$ by maximizing the lower bound
\begin{equation}
LB = \sum_{j=1}^M \sum_{i=1}^K \eta_{ij} \log( \lambda_i \mathcal{N}(x_{jk};\mu_i, \Sigma_i) )
\end{equation}
To maximize this expression, we can take derivatives with respect to the variables of interest and obtain
\begin{equation} \label{eq:lowerbound}
\begin{split}
 \frac{\partial LB}{\partial \mu_i} &= \frac{1}{2} \sum_{j=1}^M  \Sigma_i^{-1} (x_j - u_i) \eta_{ij}   
 \\
  \frac{\partial LB}{\partial \Sigma_i^{-1}} &= \frac{1}{2} \sum_{j=1}^M \eta_{ij}  \left( \Sigma_i -   (x_j-\mu_i)(x_j-\mu_i)^T \right)
\end{split}
\end{equation}
Now we can integrate these three expressions over the two dimensional surface of a triangle via a change of variables substitution, then set the result  equal to zero and solve, thus obtaining the update equations. For clarity we will also define a weight variable $w_{ij}$ and its corresponding normalization constant $W_i$
$$ w_{ij} = \eta_{ij} \alpha_j $$
$$ W_i = \sum_{j=1}^M w_{ij}$$

The resulting lower-bound likelihood can be written as
\begin{equation}
\label{eq:summary1}
\begin{aligned}
\log(L) &\geq  \frac{1}{2} \sum_{j=1}^{M}  \sum_{i=1}^{K} w_{ij} \\
& \left[2 \log(\lambda_i) - k\log(2 \pi) - \log(\det(\Sigma_i)) \right. \\
&- (\mu_j-\mu_i)^T \Sigma_i^{-1} (\mu_j-\mu_i)\\
&- \frac{1}{12} (A_j^T \Sigma_i^{-1} A_j + B_j^T \Sigma_i^{-1} B_j + C_j^T \Sigma_i^{-1} C_j \\
&\left.-3 \mu_j^T \Sigma_i^{-1} \mu_j) \right]
\end{aligned}
\end{equation}
The new mean is obtained as simply weighted mean of centroids. This update equation is identical to the one derived for the approximation in equation~\ref{eq:approx1} 
\begin{dmath}\label{eq:summary2}
 \mu_i = \frac{1}{W_i} \sum_{j=1}^M  w_{ij} \mu_j
 \end{dmath}
The same technique will provide an answer to the update equation for covariance. 
 \begin{dmath}\label{eq:summary3}
 \Sigma_i =\frac{1}{W_i} \sum_{j=1}^M  w_{ij}  \left[    \underbrace{(\mu_j-\mu_i)(\mu_j-\mu_i)^T}_{cov(\mu_j,\mu_i)}    + \underbrace{\frac{1}{12}(A_j A_j^T + B_j B_j^T + C_j C_j^T - 3 \mu_j \mu_j^T)}_{cov(\triangle_j)}  \right] 
 \end{dmath}
The final update is surprisingly simple, it is the area weighted average of the covariance obtained by using centroids as point measurements plus the covariance equation for a triangle. That is, at every iteration, each mixture is updated with some fraction of the structure of the triangles associated with it. A visual example of ellipsoids showing triangle covariance structures is shown in Fig.~\ref{fig:head}. Our derived expression for the covariance of a triangle is expressed in terms of vertices, but is consistent with the standard formulation in CGAL~\cite{gupta:inria-00327027}.\footnote{The reference document has a typographic error in the moment matrix for triangles, which should be a 5x multiple of the one for 3D tetrahedrons. The CGAL source code correctly implements this matrix in practice.} The update equation for eq.~\ref{eq:approx1} is similar, simply lacking the $cov(\triangle_j)$ term.

\subsection{Evaluating the derived loss function}
\begin{figure}[t]
\centering
\includegraphics[width=\linewidth]{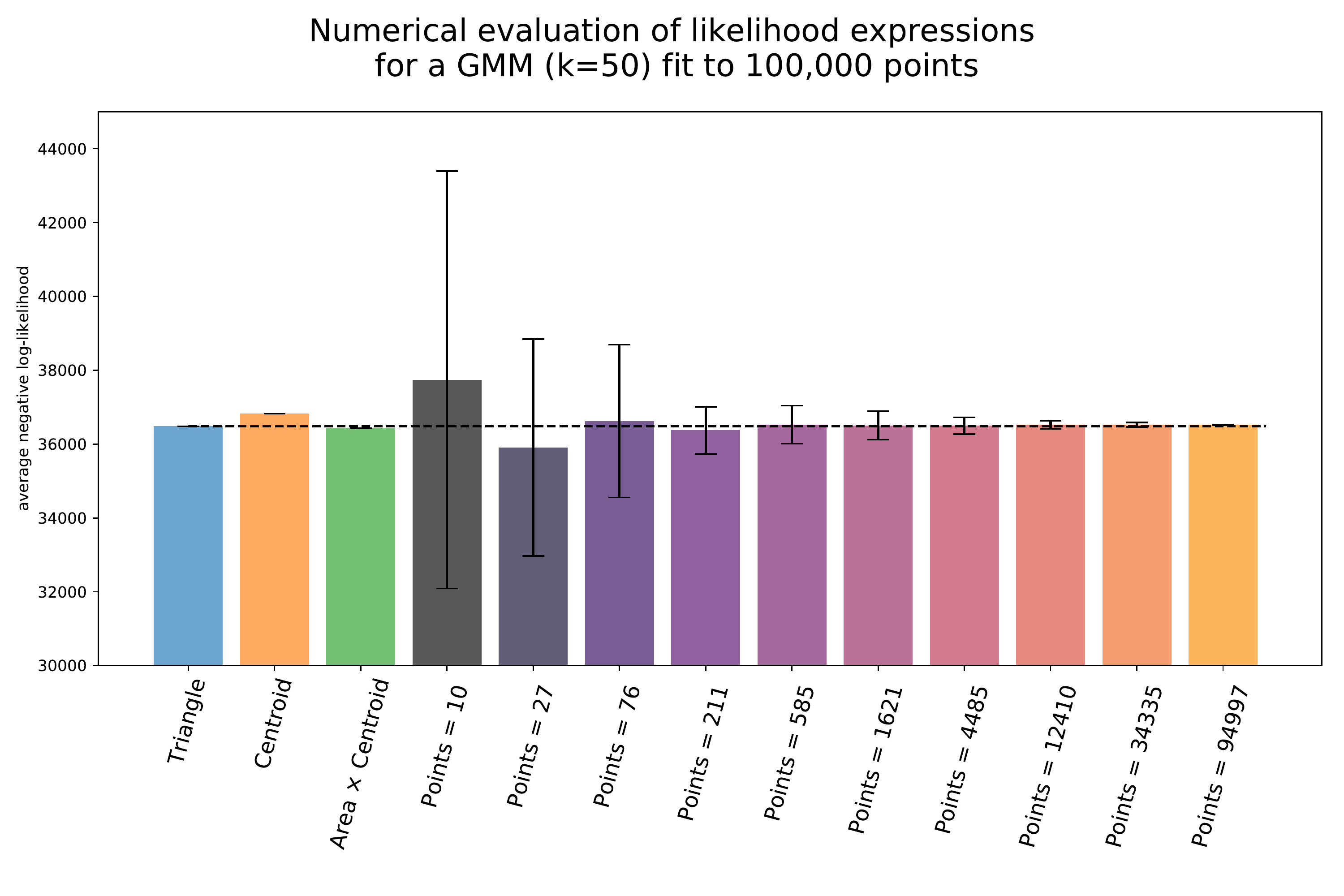}
\caption{\label{fig:llgraph} \textbf{Comparison of fitting metrics}. 
After fitting a 50 mixture GMM to 100,000 points randomly sampled from the \textit{res4} Stanford bunny, different log-likelihood expressions are compared. \textit{Triangle} refers to the equation~\ref{eq:summary1}, while \textit{Centroid} refers to using only the triangle centroids, while \textit{Area $\times$ Centroid} refers to using our approximation in eq.~\ref{eq:approx1}. The results are on the \textit{res4} variant of the Stanford bunny, which has 948 faces and 453 vertices. The remaining bars show results using different numbers of points sampled from the mesh surface. The y-axis shows the sum of individual Gaussian component log-likelihoods ($\sum \sum \log(x) $), equivalent to the lower-bound obtained from Jensen's inequality. The horizontal line shows the result of using all the points, our best approximation of the correct answer. }
\end{figure}

To evaluate the validity of the expression in equation~\ref{eq:summary1}, we compare its fitting fidelity numerically against a large number of sampled points. The results are shown in figure~\ref{fig:llgraph}. We can see that, the lower bound expression for triangles is equal to that obtained numerically from a large number of points. Since the lower-bound expression is all that's needed in EM optimization~\cite{Dempster1977}, the equality of this expression suggests we can use it in fitting real data.

\section{Results}\label{sec:results}
We performed experiments fitting Gaussian Mixture Models to triangular meshes. We swept a wide range of $K$ mixture components (from 6 to 400) and evaluated  two different initialization schemes. The first initialization performs k-means++~\cite{arthur2007k} clustering, and uses those clusters as initial assignments for the EM method. The second method uses simple random assignments for initialization. We run 25 iterations of EM for all methods, with a tight tolerance ($\epsilon=10^{-12}$) to prevent an early exit from the optimization. 

To evaluate the converged model, we use a densely sampled point cloud of the initial mesh (Figure~\ref{fig:examplepts}) and report its likelihood according to equation~\ref{eq:pdf}. Since all of our experiments tend to operate on 1,000 points or triangles, the use of 50,000 points for evaluating the model should provide a good test of GMM model fidelity. In all of these cases, we focus on the Stanford Bunny. Visual examples of our input and output data is shown in Figure~\ref{fig:examplepics}. 

\subsection{Mesh Input Data} \label{sec:trivert}
\begin{figure}
\centering
\includegraphics[width=\linewidth]{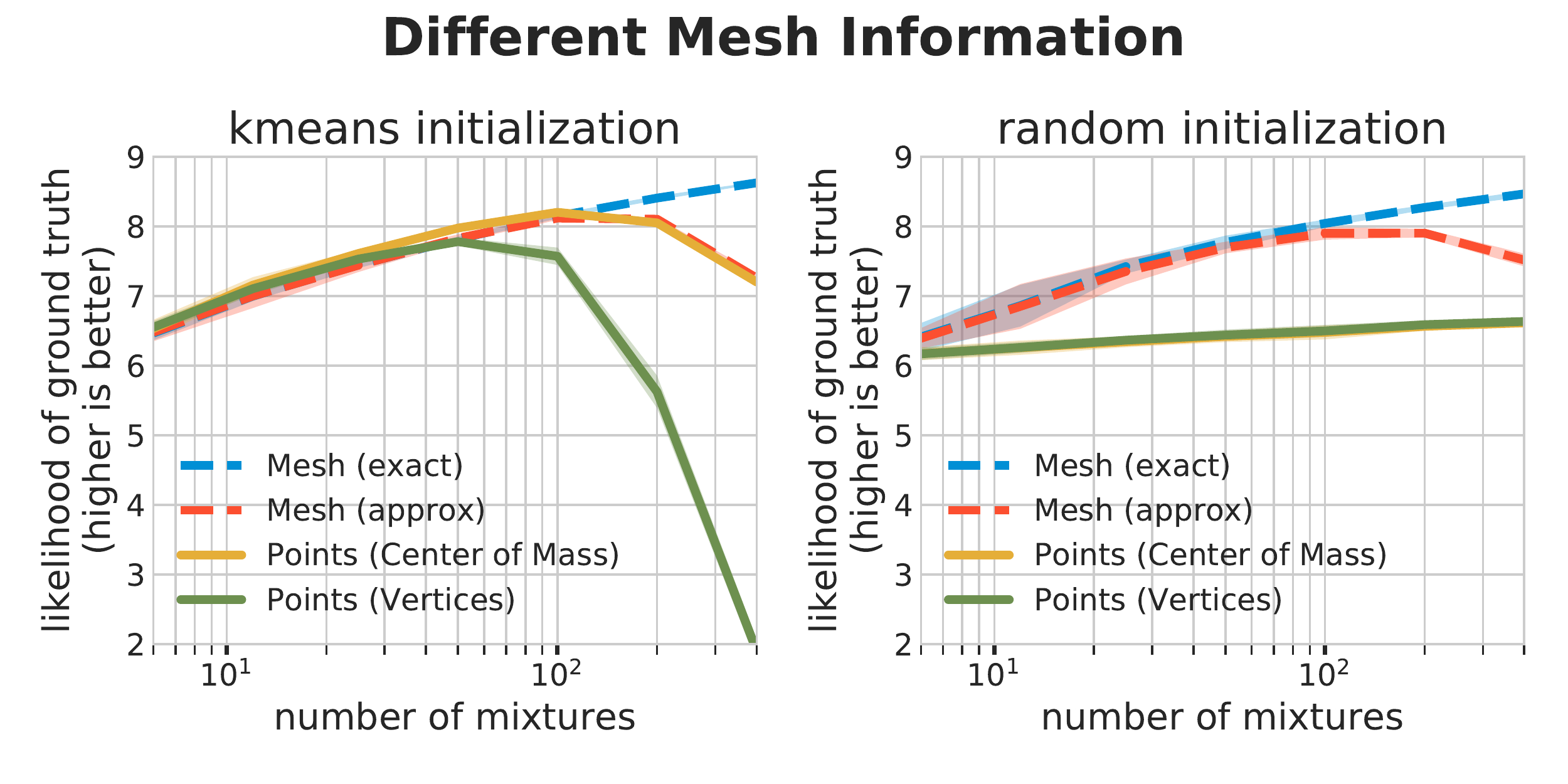}
\caption{\label{fig:graph2} \textbf{GMMs fit to different data-types of the low-res Stanford Bunny}. The graphs show fitting fidelity of the converged model. The dashed lines use triangle likelihood estimates, while the solid lines use traditional point loss. \textit{Exact} refers to the M-step derived in eq.~\ref{eq:summary2} \& \ref{eq:summary3}, while \textit{Approx} refers to only using eq.~\ref{eq:approx1}. The results are on the res4 variant of the Stanford
bunny, which has 948 faces and 453 vertices. Evaluation is performed by evaluating the likelihood of 50,000 points sampled from the \textit{res4} Stanford bunny. }
\end{figure}

The first set of experiments, shown in Figure~\ref{fig:graph2}, compares fitting GMM models to different input formats of the \textit{res4} Stanford Bunny. We compare our exact (eq. \ref{eq:summary1}) and approximate (eq. \ref{eq:approx1}) mesh loss equations against fitting a traditional point-loss to the triangle centroids and the mesh vertices. The best results came from our exact mesh expression, with the second best being its approximation. The proposed methods handled random initialization and k-means initialization. On the other hand, the point-based methods often had a preferred initialization. A qualitative look at the resulting models, shows that the mesh GMM (Fig.~\ref{fig:meshgmm}) produces a fuller model of the Stanford Bunny than the center-of-mass GMM (Fig.~\ref{fig:ptsgmm}), even when the evaluated likelihood was numerically very similar.

\subsection{Mesh Decimation}

While the above experiments used the low-resolution \textit{res4} Stanford Bunny, we also evaluated which GMM fitting strategy works best when subsampling high-resolution mesh data. In our case, we try two methods point-sampling (Poisson and Random) and two methods of triangle collapse (Quadric and Clustering).  Our experimental procedure involves fitting a GMM to a low-resolution mesh (1,000 faces) or point cloud (1,000 points) and evaluating the likelihood of the resulting GMM against a dense point sampling of the original shape (50,000 points). The different sampling strategies can be seen visually in Figure~\ref{fig:examplepics}. 

To generate a low-resolution mesh, we try two methods, clustering decimation~\cite{Rossignac1993}, which is fast, and quadric error decimation~\cite{Garland1997}, which is more accurate. To generate a low-resolution point cloud, we both randomly sample points on the surface on the mesh, and use Poisson Disc sampling to ensure uniform samples~\cite{Cignoni2012}. As before, the first is faster while the latter produces better results. Poisson Disc sampling is also used to generate the high-resolution, "ground truth", point cloud used for evaluation.

\begin{figure}
\centering
\includegraphics[width=\linewidth]{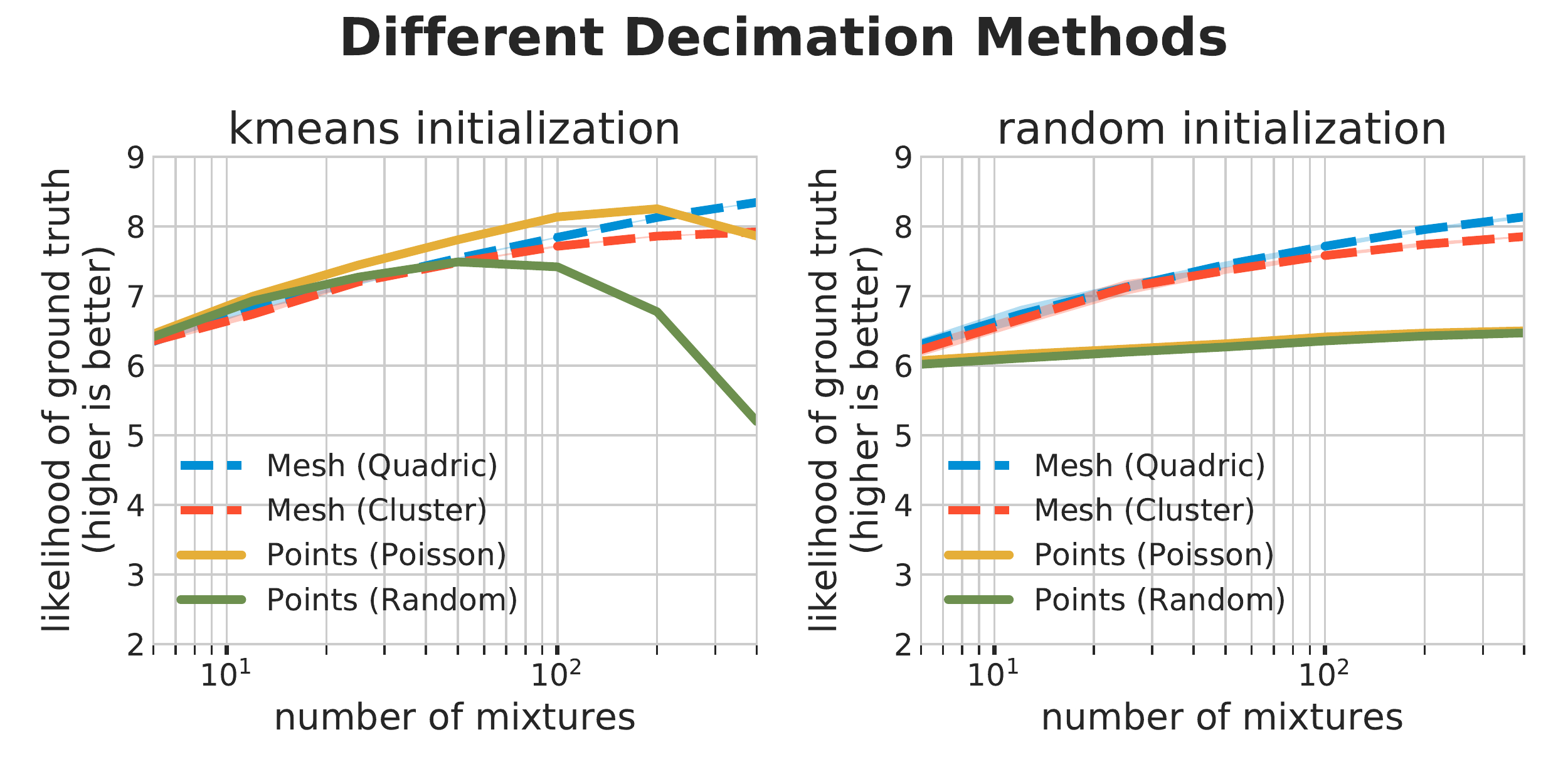}
\caption{\label{fig:graph} \textbf{GMMs fit to different input formats of the Stanford Bunny}. The  graphs show fitting fidelity of the convered Gaussian Mixture MModel. The dashed lines use eq.~\ref{eq:summary2},\ref{eq:summary3}, while the solid lines use traditional point loss. The results are on the Stanford bunny, which has been simplified to $\approx 1000$ triangles or points respectively with two different methods each. Random or Poisson disc sampling ~\cite{Cignoni2012}, and with either clustering~\cite{Rossignac1993} or quadric-error decimation~\cite{Garland1997}. See Fig.~\ref{fig:examplepics} for visual examples of these formats. Evaluation is performed by evaluating the likelihood of a high-resolution point cloud sampled from the original Stanford bunny. }
\end{figure}

\begin{figure*}
    \centering
    \begin{subfigure}[b]{0.23\textwidth}
        \includegraphics[width=\textwidth]{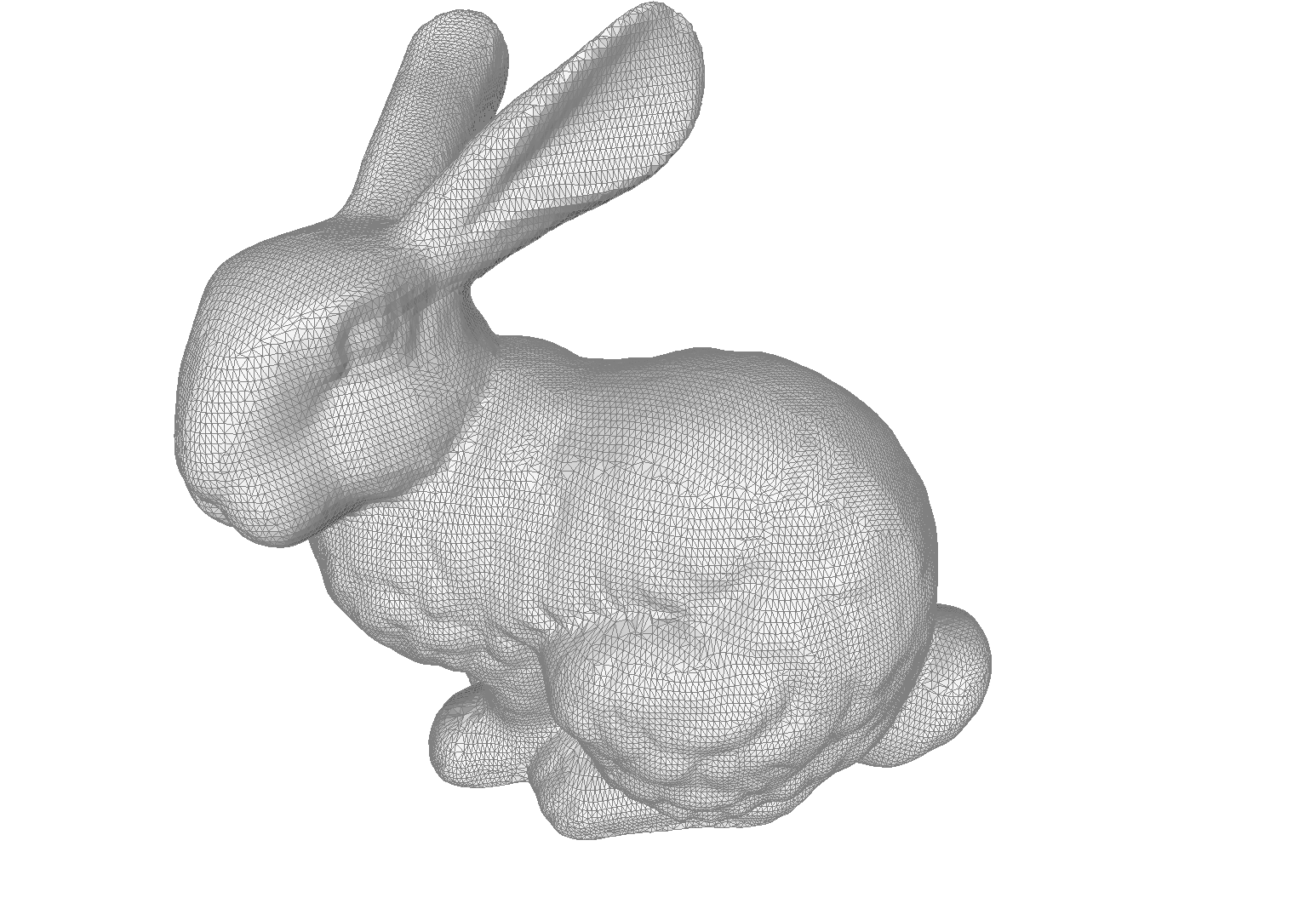}
        \caption{Stanford Bunny}
        \label{fig:bf}
    \end{subfigure}
    ~ 
    \begin{subfigure}[b]{0.23\textwidth}
        \includegraphics[width=\textwidth]{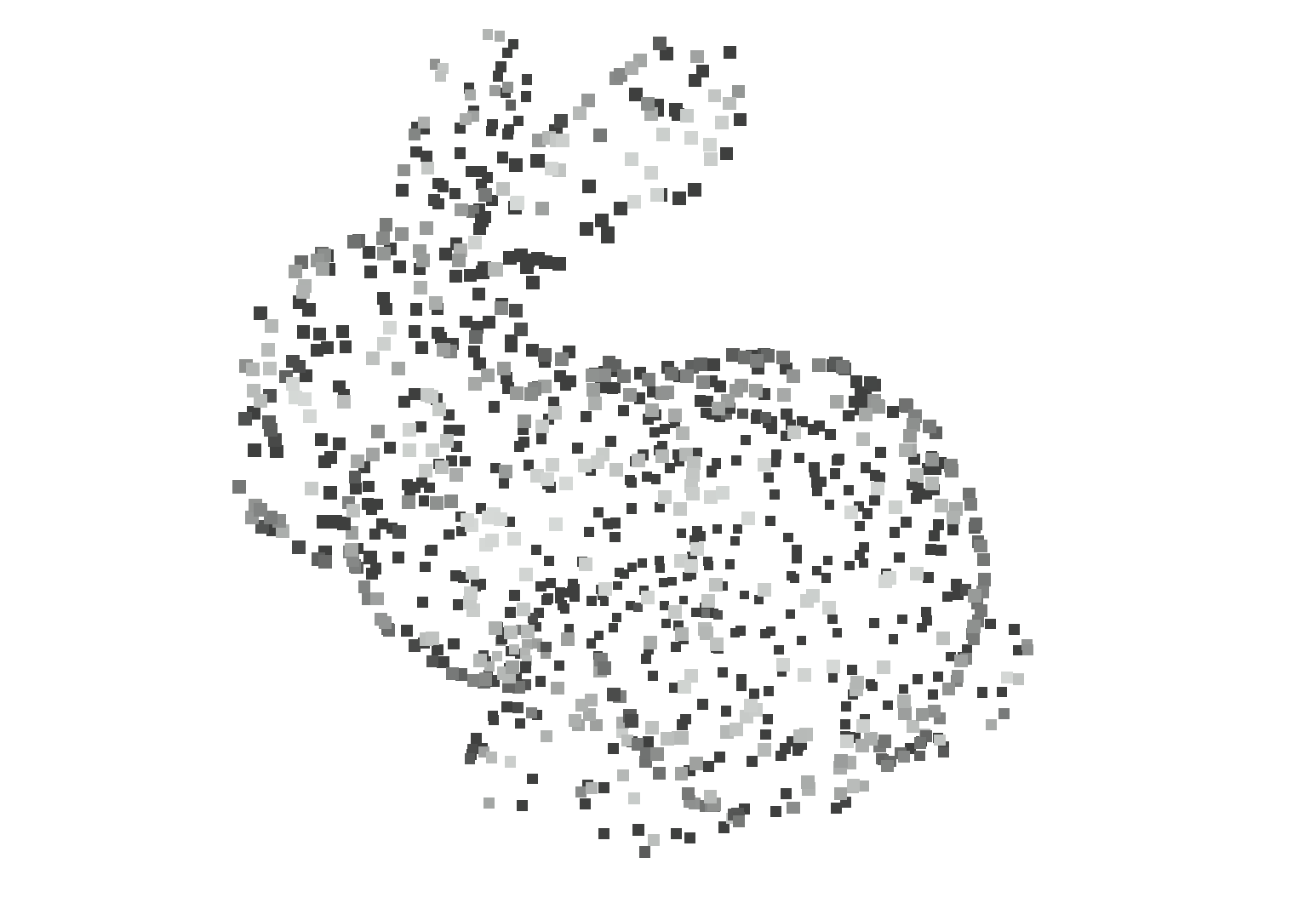}
        \caption{Random Samples}
        \label{fig:random}
    \end{subfigure}
    ~ 
    \begin{subfigure}[b]{0.23\textwidth}
        \includegraphics[width=\textwidth]{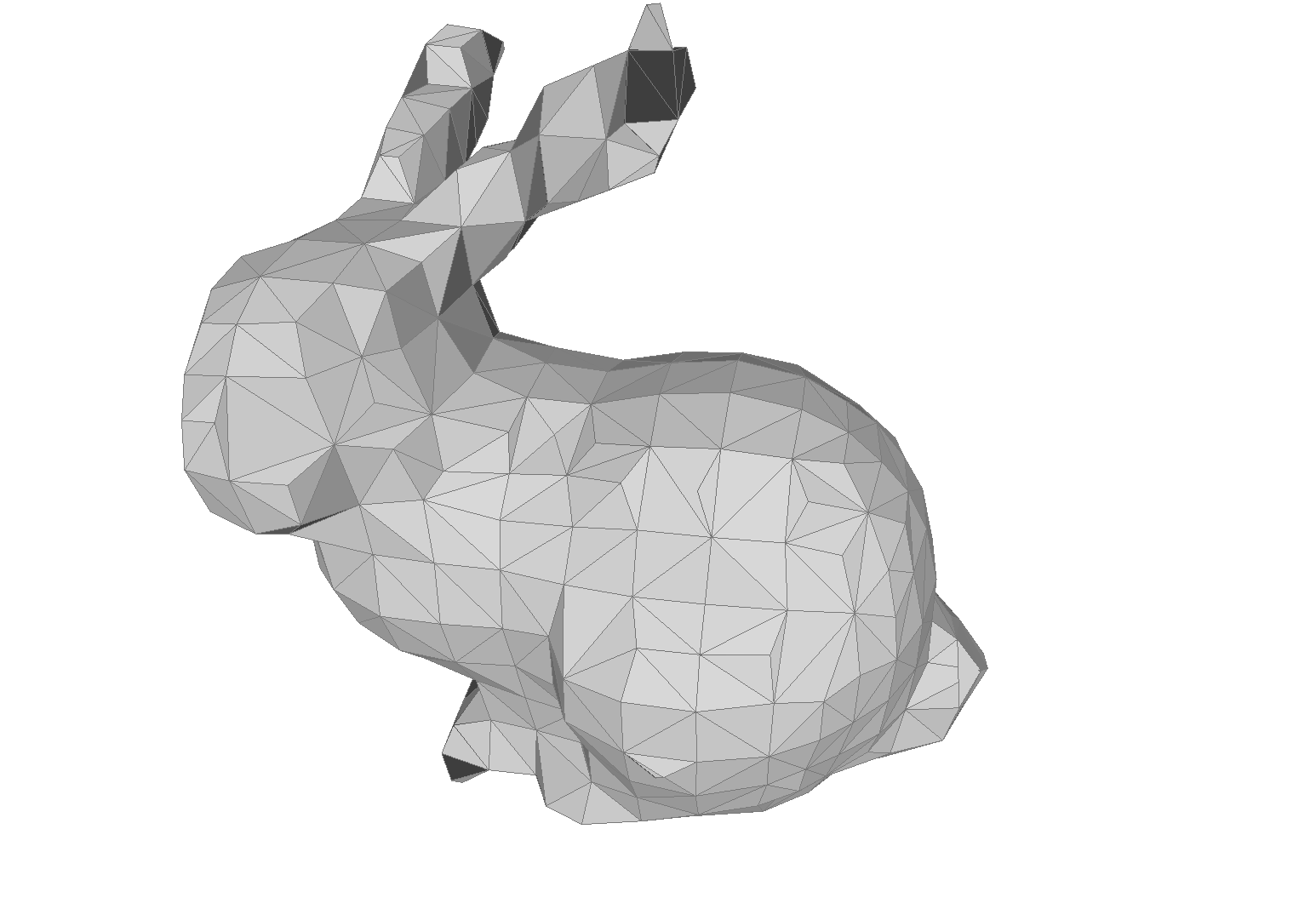}
        \caption{Clustering Decimated~\cite{Rossignac1993}}
        \label{fig:mouse}
    \end{subfigure}
    \begin{subfigure}[b]{0.23\textwidth}
        \includegraphics[width=\textwidth]{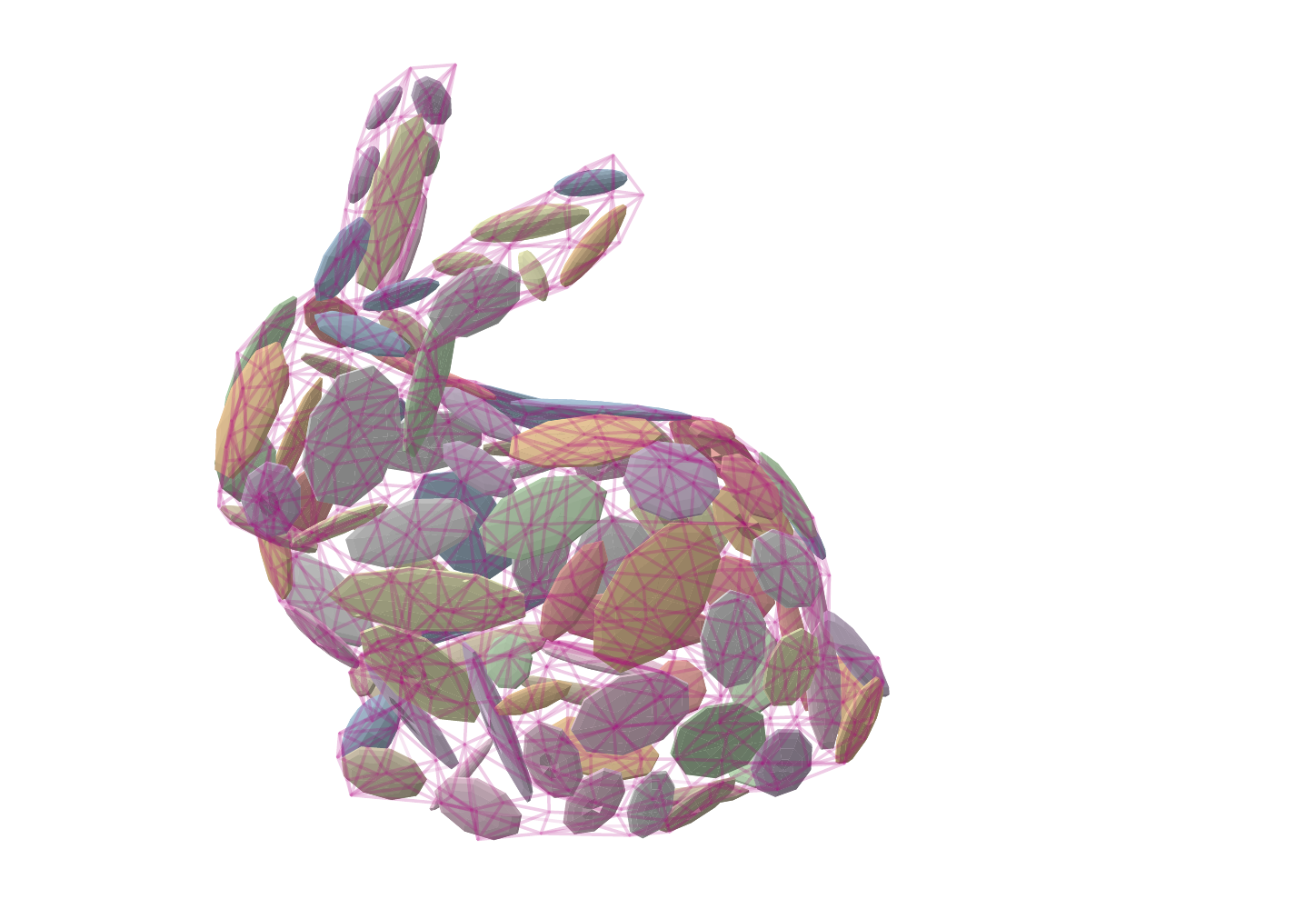}
        \caption{Points GMM ($K=100$)}
    \label{fig:ptsgmm}
    \end{subfigure}
        \begin{subfigure}[b]{0.23\textwidth}
        \includegraphics[width=\textwidth]{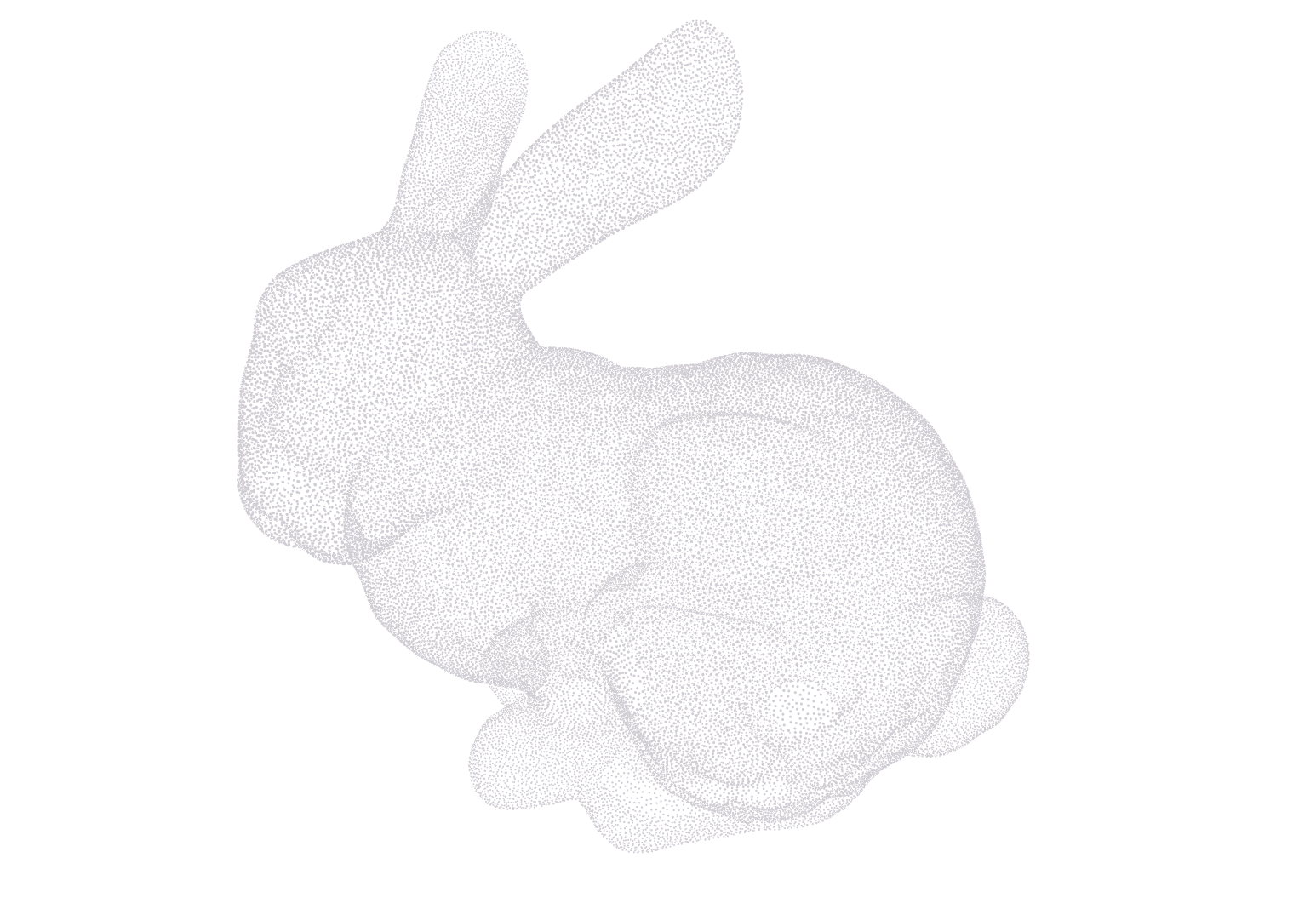}
        \caption{Dense Sampled}
        \label{fig:examplepts}
    \end{subfigure}
    ~ 
    \begin{subfigure}[b]{0.23\textwidth}
        \includegraphics[width=\textwidth]{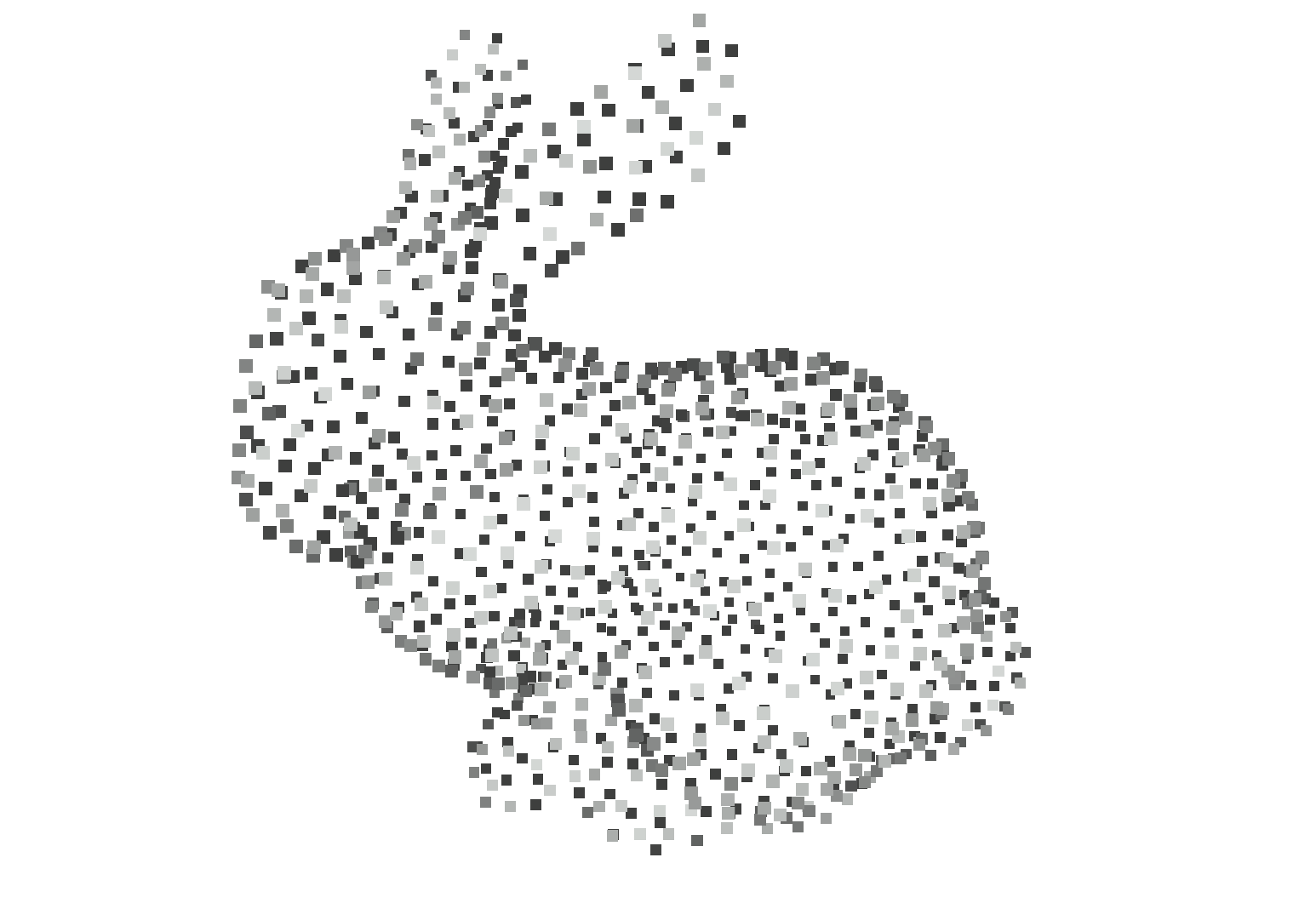}
        \caption{Poisson Disc Sampled~\cite{Cignoni2012}}
        \label{fig:poisson}
    \end{subfigure}
    ~ 
    \begin{subfigure}[b]{0.23\textwidth}
        \includegraphics[width=\textwidth]{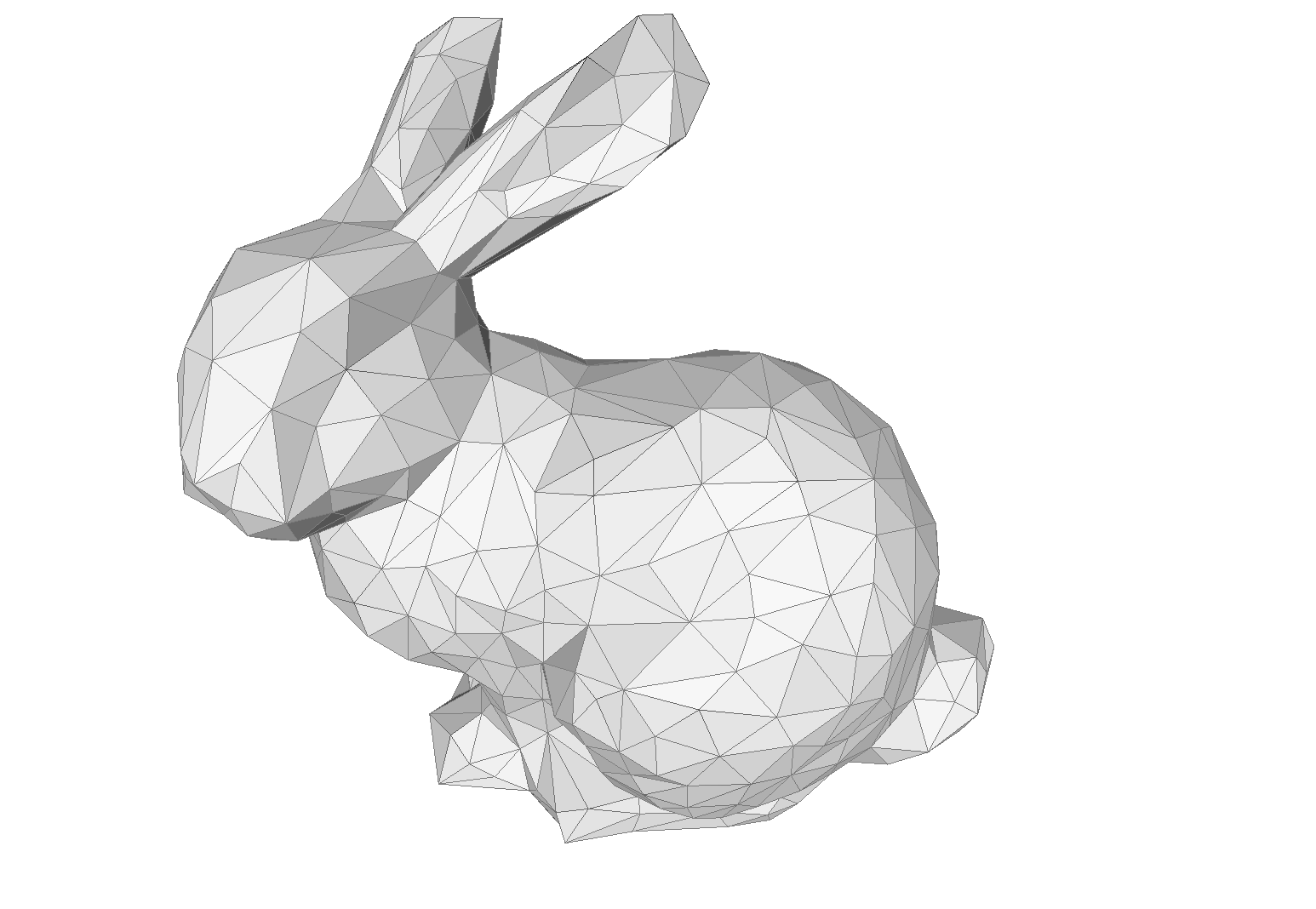}
        \caption{Quadric Decimated~\cite{Garland1997}}
        \label{fig:quadric}
    \end{subfigure}
    \begin{subfigure}[b]{0.23\textwidth}
        \includegraphics[width=\textwidth]{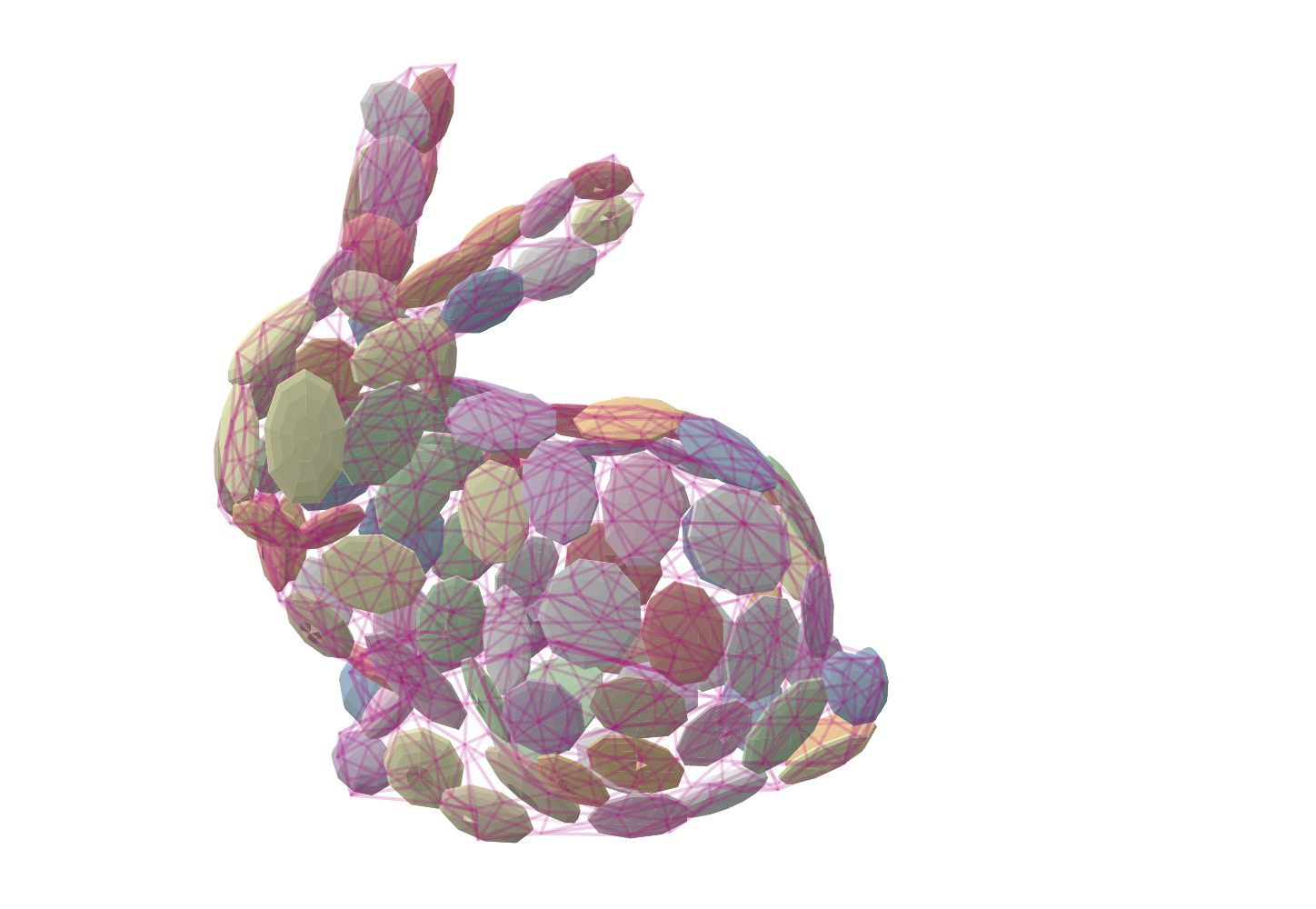}
        \caption{Mesh GMM ($K=100$)}
    \label{fig:meshgmm}
    \end{subfigure}
    \caption{Examples of different input and output representations for the Stanford Bunny}\label{fig:examplepics}
\end{figure*}

The results of these experiments are reported in Figure~\ref{fig:graph}. As before, the mesh-based registration strategies are largely invariant to initialization method while the point-based strategies often prefer k-means initialization (exceptions discussed in Sec.~\ref{sec:discu}). The best results often came from the use of Poisson Disc Sampling (with k-means), which ensures uniform coverage of the surface areas. In contrast, using random samples generated the lowest quality results. The mesh-based techniques proved to be reliable across all tests, regardless of mixture number and initialization.
\subsection{Discussion}\label{sec:discu}
All of these experiments were run using the \textit{GaussianMixture} code base from \texttt{scikit-learn v0.20.0} ~\cite{scikit-learn}. We modified the code to support additional weight and covariance terms (Sec.~\ref{sec:general}), which were general enough to allow us to implement all proposed methods. 

One surprising result in Figures ~\ref{fig:graph2} and~\ref{fig:graph} was that random sampled points performed better with random initialization than k-means initialization (at high mixture numbers). As the EM algorithm only finds a local minima, this suggests that k-means may not always be an ideal initialization technique. We believe that this occurred due to either a bad local minima from initialization, or fairly flat cost during optimization, leading to an early exit condition being triggered (despite our tight tolerance of $\epsilon = 10^{-12}$). This behavior was never observed when using our proposed exact mesh formulation. 

We used 25 iterations for all of these experiments. When 100 iterations were used, the point-based methods performed better (nearly as good as our proposed method). However, our method often converged in about $\frac{1}{3}$ the number of iterations, so we picked a lower iteration number for consistency in runtime.

\section{Extensions}
\subsection{Generalization to other primitives}\label{sec:general}
While equations \ref{eq:summary1},\ref{eq:summary2},\ref{eq:summary3} were derived specifically for triangles, the update equations can be written more generally for any primitive (triangles, Gaussian mixtures, cuboids, etc.) using $\mu_p,\Sigma_p,\alpha_p$ to denote primitive's mean, covariance and size ($\int_S dS $) respectively. Then the loss, mean update, and covariance update equations for a Gaussian Mixture can be written with equations \ref{eq:general1},\ref{eq:general4},\ref{eq:general6}. The previous equations can be seen as a special case of these formulas, which provide an M-step update for any set of geometric primitives $p \in \textbf{P}$ in fitting a Gaussian Mixture Model. 
\begin{align}
\log(L) &\geq \frac{1}{2} \sum_{p=1}^{P} \sum_{i=1}^{K} w_{ip} \left[  2 \log(\lambda_i) - k\log(2 \pi) \label{eq:general1} \right. \\
&- \left. \log(\det(\Sigma_i)) -  (\mu_p-\mu_i)^T \Sigma_i^{-1} (\mu_p-\mu_i) - \Sigma_p \right] \nonumber
\end{align}
\begin{align}
 \mu_i &=  \frac{1}{W_i} \sum_{p}^P w_{ip} \mu_p \label{eq:general4} \\ 
 \Sigma_i &= \frac{1}{W_i} \sum_{p}^P w_{ip}  \left[ (\mu_p-\mu_i)(\mu_p-\mu_i)^T   + \Sigma_p  \right]  \label{eq:general6}
\end{align}
We note that these are the exact same update equations used in fitting hierarchical Gaussian Mixture Models~\cite{NIPS1998_1543}. However, while previous work  applied these equations to fitting GMMs to existing GMMs, we show how this update can be used for fitting geometric data. This general form allows for easy substitution of known structural information (in the form of a second moment) about any geometric primitive. Computing covariance structures for arbitrary polyhedra is well studied area of research~\cite{fastCovariance,gupta:inria-00327027}. 
 
\subsection{Number of Mixtures}
In our later experiments, we fix the number of mixture models. Unless otherwise stated, we use $K=100$. A visual example of this mixture can be see in Figures~\ref{fig:ptsgmm} and~\ref{fig:meshgmm}. We picked $K=100$ as this matches the experimental conditions recommended for using GMMs for SLAM~\cite{wenniegmm}. In practice, there are many ways to select this number, including flatness of the distribution's KL-divergence ~\cite{shobitGMMNum,dhawale2018fast}, flatness of the mixture themselves~\cite{Eckart2018}, or by evaluating an information criterion~\cite{Pelleg00x-means:extending}. We believe that when this technique is used in practice, this number can either be found through cross-validation~\cite{wenniegmm} on a registration dataset or by using external system information such as depth sensor noise models~\cite{Keselman_2017_CVPR_Workshops}. 

\begin{figure*}[ht!]
\centering
\includegraphics[width=\linewidth]{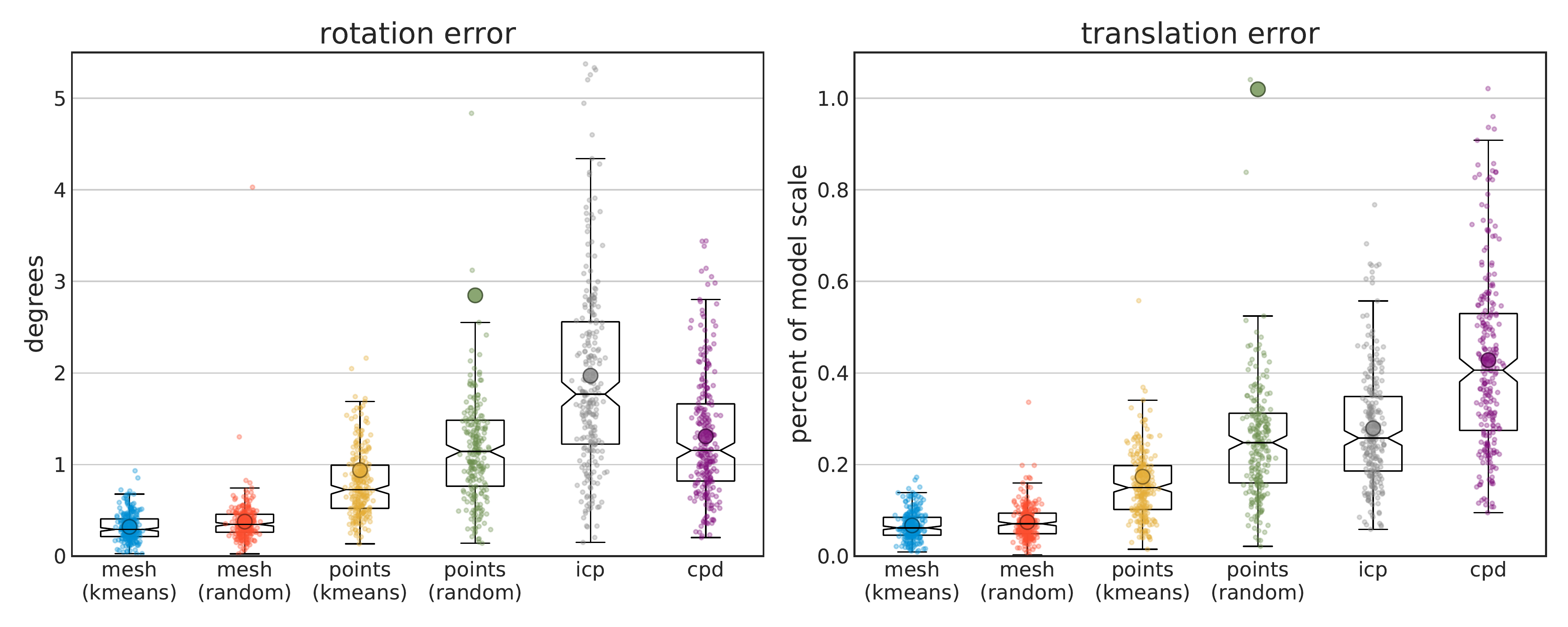}
\caption{Registration results on the Stanford Bunny, following the experimental setup described in in~\cite{Eckart2018}. The experimental conditions for these tests are described in Section~\ref{sec:applications}. We plot the results of 250 random rigid deformations. We show the actual data, along with a box and whisker plot showing the median and its confidence interval; the  mean is plotted as a larger dot. \textit{mesh} and \textit{points} are the results of maximizing the likelihood of a set of points (N=453) against our fit Gaussian Mixture models (K=100). The mesh result uses our proposed method, fitting a GMM to the mesh triangles, while points shows the results of fitting a GMM to the mesh vertices (N=453). \textit{icp} is our implementation of point-to-point ICP~\cite{mckaybesl}, and \textit{cpd}~\cite{cpd} is from pyCPD~\cite{PyCPD}. \textit{Model size} refers to the length of the diagonal of the model's bounding box, and we report our results in percent (so all reported methods have position error, on average, better than 1.1\% of the model size). Some methods have outliers that converged to the wrong local minima, and hence have a very large mean relative to their distribution. }
\label{fig:regtest}
\end{figure*}

\section{Applications}\label{sec:applications}
The experiments in section~\ref{sec:results} showed that fitting Gaussian Mixture Models using structural information tends to produce higher quality probability distributions. Some recent work has focused solely on the efficient nature of GMMs in representing shapes~\cite{Eckart2016}. Here we show that our improvements in model quality produce an appreciable performance improvement in actual 3D computer vision applications.

Gaussian Mixture Models have found wide use in the 3D registration literature. From the Normal Distance Transform~\cite{5152538,doi:10.1177/0278364912460895}, to variants of the $L_2$ loss~\cite{pointsetregistration,L2LossNDT3D} and even Coherent Point Drift~\cite{cpd}, many 3D registration methods utilize Gaussian Mixture Models. Their benefits include robustness to noise, smooth variation over 3D space, speed of evaluation, and straightforward control over model complexity. These models can provide results that are state-of-the-art in both runtime and registration accuracy~\cite{Eckart2018}. We show that applying our proposed mesh GMM fitting can produce an improvement in these results. 

\subsection{Mesh Registration}
We replicate the experimental setup of a recent paper~\cite{Eckart2018}, demonstrating how Gaussian Mixture Models can be used for efficient 3D registration. As our experimental setup matches~\cite{Eckart2018}, the 20 different dozen registration methods compared in Figure 3 of that work can be directly compared against the results here. Their experiment operates on taking a large number of random deformations of the Stanford Bunny and evaluating the final quality of fitting result. Our results can be seen in Figure~\ref{fig:regtest}.

To perform 3D registration, we first build a GMM for the \textit{res4} Stanford Bunny, as in Section~\ref{sec:trivert}, using 100 iterations of EM with a tolerance of $10^{-5}$. We then sample vertex number of 3D points from the surface of the mesh ($N=453$). While previous work has focused on a Point-to-Distribution (P2D) technique with a polynomial approximation of the likelihood function~\cite{Magnusson_2009}, we do straightforward P2D in our experiments. Our registration process consists of finding the rigid body transformation that maximizes equation~\ref{eq:pdf}. We use the identity transformation for initialization and then perform gradient-based optimization to find the local minima. We compute gradients using numerical differences. For the optimizer, we tried both Conjugate Gradients~\cite{cgog,Shewchuk94anintroduction} and BFGS~\cite{NoceWrig06} as optimization strategies, which produced similar results and we report the BFGS results as it ran faster. 

To parameterize our rigid-body transformation, we perform the optimization on $\mathbb{R}^7$, with a translation $t \in \mathbb{R}^3$ and a quaternion $q \in \mathbb{R}^4$.  Quaternions are well studied in the context of optimization for rigid-body transformation~\cite{Schmidt:2001:UQP:647260.718651,Hartley2009}. As we use numerical differences in our optimizer, we did not utilize methods the closed-form gradients for quaternions~\cite{quaterniongrads}. While many authors prefer the exponential map for optimizing rigid-body transformations~\cite{wenniegmm}, our experiments using the rotation vector $v = \theta \hat{v} \in \mathbb{R}^3$ (with rotation angle $\theta$ around the unit vector $\hat{v}$) produced nearly identical results in our final registration result. 

We performed these experiments on our mesh and point-derived Gaussian Mixture Models, as well as two baselines. We implemented our own point-to-point Iterative Closest Point (ICP) method~\cite{mckaybesl} and used an exiting implementation of Coherent Point Drift (CPD)~\cite{cpd} from pyCPD~\cite{PyCPD}. We adjusted pyCPD to run for 150 iterations to approximately match the run-time of our P2D GMM registration. ICP we ran for up to 50,000 iterations, or until the improvement in mean matching error was below $10^{-9}$. For consistency, all methods used in this paper were implemented in the Python programming language and only used the CPU.


\subsection{Analysis of Mesh Registration}
The results in Figure~\ref{fig:regtest} demonstrate that our mesh-derived Gaussian Mixture Model provides improved registration results when using P2D compared to the existing baselines, ICP and CPD. Not only does our method produce better registration on average, but it also demonstrates a better distribution of errors. Specifically, the small difference between the median and mean errors shows that our method is less prone to outliers. On the other hand, the point-based GMM P2D registration results had outliers that dragged the mean towards the worst quartile of results. The randomly initialized point-based GMM had a mean that was about three times that of its median result, suggesting that some experiments produced results in the incorrect local minima.
\subsection{Other 3D Models}\label{sec:other}
We report results on additional 3D models in Table~\ref{tab:my-label}. For consistency, we decimated each model to 1000 faces using~\cite{Garland1997} and then repeated our previous experiments exactly (except that the registration results are now the average of 25 runs). The likelihood column reports the per-sample average log-likelihood of ground truth, where larger numbers are better. The translation and rotation errors are reported as a percentage of the error obtained by ICP registration. In all our experiments, the mesh-based GMM always outperformed the point-based one, often significantly. 
\renewcommand{\arraystretch}{1.2}
\begin{table}[hb]
\centering
\begin{tabular}{@{}lrrrrrr@{}}
\toprule
\textbf{Model} & \multicolumn{2}{C{1.9cm}}{\textbf{Likelihood} \newline (larger is better)} & \multicolumn{2}{C{1.4cm}}{\textbf{Translation Error} \newline (\% of ICP)} & \multicolumn{2}{C{1.4cm}}{\textbf{Rotation Error}\newline  (\% of ICP)} \\
               & points              & mesh              & points                  & mesh                 & points                & mesh                \\ \midrule
Armadillo      & -14.6               & -12.2             & 127                     & 37                   & 161                   & 33                  \\
Bunny          & 7.6                 & 8.2               & 50                      & 28                   & 41                    & 17                  \\
Dragon         & 6.9                 & 7.6               & 68                      & 25                   & 40                    & 19                  \\
Happy          & 7.3                 & 8.2               & 101                     & 27                   & 85                    & 27                  \\
Lucy           & -21.5               & -18.3             & 95                      & 23                   & 122                   & 35                  \\ \bottomrule
\end{tabular}
\caption{Results of repeating the experiments in Figures~\ref{fig:graph2} and \ref{fig:regtest} on mulitple models from~\cite{StanfordScanRep}. All experiments used k-means initialization and $K=100$. Details in Sec.~\ref{sec:other}.}
\label{tab:my-label}
\end{table}
\subsection{Visual Odometry} \label{sec:slam}
Our proposed method can be also used for improved models of partial view observations, such as those seen in simultaneous localization and mapping (SLAM). In this case, the geometric primitives used represent not the exact surface, as with our mesh experiment above, but instead an uncertainty region for each 3D measurement.

\begin{figure}[t]
\centering
\includegraphics[width=0.8\linewidth]{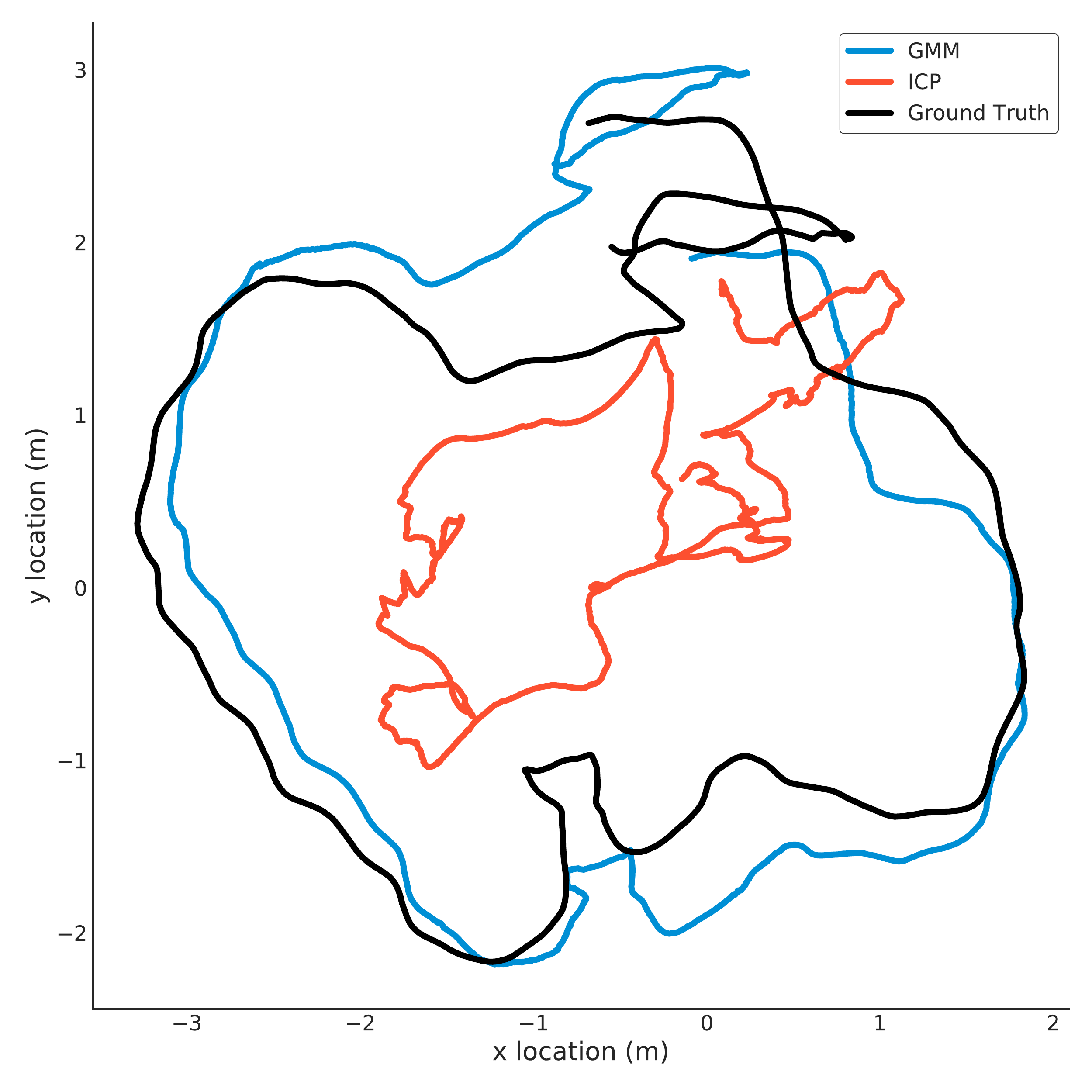}
\caption{\label{fig:slamgraph} Top-down view of trajectories generated using different registration methods for visual odometry. The GMM method uses a per-pixel uncertainty primitive during GMM fitting. The ICP method is pt2pt ICP~\cite{mckaybesl}. Runtime for both methods was similar. For details, see Section~\ref{sec:slam}. For additional baselines and experiments on this dataset see~\cite{wenniegmm}. }
\end{figure}

We performed experiments using a GMM distribution-to-distribution (D2D) registration method for visual odometry~\cite{wenniegmm}, reproducing an experiment on an RGB-D dataset sequence from the TUM dataset (\texttt{freiburg3 long office household})~\cite{sturm12iros}. We are able to incorporate structural information into the fitting of the Gaussian Mixture Models by adding depth uncertainty information around each 3D point and applying equation~\ref{eq:general6} during GMM fitting. The results are shown in Figure~\ref{fig:slamgraph}. 

For our registration experiments, we first subsampled the depth images to $160 \times 120$ resolution before performing frame-to-frame registration over the 2510 frame sequence. The ICP method used our aforementioned point-to-point ICP method over 2,500 points randomly sampled from each point cloud  (selected to roughly match the run-time performance of our method). The GMM method fit a $K=100$ GMM to the point cloud using our uncertainty model, and performed registration using a D2D metric~\cite{wenniegmm}. We used the determinant-free method as it was much faster in our re-implementation. Additionally, our implementation used numerical gradients and BFGS~\cite{NoceWrig06} as the optimizer. 

In this case, our primitive model was simple one, a rectangle representing the size of each 3D measurement in the X and Y axes of the camera. This generated a trajectory with absolute translational error RMSE of $0.878 m$, a small improvement ($2.4\%$) over the $0.899 m$ APE produced from building GMMs without uncertainty primitives. Additionally, we found that D2D registration time was $22\%$ faster when using GMMs built from primitives. 

\section{Conclusion}\label{sec:conclusion}
We have shown how to build Gaussian Mixture Models by incorporating structural information into their Expectation Maximization algorithm. We demonstrate theoretical and empirical equivalence with traditional techniques, along with providing a fast approximation to our proposed method. By using the covariance structure from the triangles of a mesh, we are able to build GMM models more quickly, robustly and to higher quality. Additionally, these models provide an improved result in 3D registration. For a theoretical understanding of how geometric structures, point samples, and integrals interact, our product integral derivation provides a model that is invariant to resampling (such as triangles being merged or split while retaining the same overall 3D structure). We believe that paper demonstrates that using structural information can lead to methods that are faster, more robust, and more lead to improved performance. 

\clearpage
\bibliographystyle{IEEEtran}
\bibliography{sample}
%

\end{document}